# Examining the physical and psychological effects of combining multimodal feedback with continuous control in prosthetic hands

Digby Chappell[1,2,*], Zeyu Yang[3], Angus B. Clark[4], Alexandre Berkovic[5,6], Colin Laganier[7], Weston Baxter[1], Fernando Bello[2], Petar Kormushev[1], and Nicolas Rojas[1]

[1]Dyson School of Design Engineering, Faculty of Engineering, Imperial College London, London, UK
[2]Department of Surgery and Cancer, Faculty of Medicine, Imperial College London, London, UK
[3]Department of Engineering Science, University of Oxford, Oxford, UK
[4]Department of Bioengineering, Faculty of Engineering, Imperial College London, London, UK
[5]MIT Sloane School of Management, Massachusetts Institute of Technology, Boston, USA
[6]MIT Operations Research Centre, Massachusetts Institute of Technology, Boston, USA
[7]Department of Computer Science, Faculty of Engineering, University College London, London, UK
[*]d.chappell19@imperial.ac.uk

**ABSTRACT**

Myoelectric prosthetic hands are typically controlled to move between discrete positions and do not provide sensory feedback to the user. In this work, we present and evaluate a closed-loop, continuous myoelectric prosthetic hand controller, that can continuously control the position of multiple degrees of freedom of a prosthesis while rendering proprioceptive feedback to the user via a haptic feedback armband. Twenty-eight participants without and ten participants with limb difference were recruited to holistically evaluate the physical and psychological effects of the controller via isolated control and sensory tasks, dexterity assessments, embodiment and task load questionnaires, and post-study interviews. The combination of proprioceptive feedback and continuous control enabled accurate positioning, to within 10% mean absolute motor position error, and grasp-force modulation, to within 20% mean absolute motor force error, and restored blindfolded object identification ability to open-loop discrete controller levels. Dexterity assessment and embodiment questionnaire results revealed no significant physical performance or psychological embodiment differences between control types, with the exception of perceived sensation, which was significantly higher ($p < 0.001$) for closed-loop controllers. Key differences between participants with and without upper limb difference were identified, including in perceived body completeness and frustration, which can inform future prosthesis development and rehabilitation.

## Introduction

Prosthetic hands, in general, exist to restore the lost function of the hand or arm of a user who has an upper limb difference (ULD), be it congenital or acquired from trauma. However, human dexterity is a sophisticated coupling between motor and sensory neural pathways, culminating in a manipulation system that dwarfs even the best robotic hands in terms of functionality[1]. Although some robotic hands approach the mechanical complexity of their human counterpart[2,3], the majority of prosthetic hands offer between 1 and 7 controllable degrees of freedom (DOFs)[4–8]. These are generally arranged to control finger flexion and, in some cases, a secondary thumb or wrist rotation. With cost, space, and weight constraints limiting design[9], it remains unlikely that prosthetic hands will approach the mechanical functionality of human hands in the near future. The challenge, therefore, is how to control the available DOFs of a prosthesis in the best way.

The most common clinically available, multi-DOF, electromechanical prosthesis control method is to use pattern recognition to classify muscle activity, recorded via electromyography (EMG), into predefined actions, without providing any sensory feedback to the user[10,11]. Improving accuracy and increasing the number of actions have been the focus of research dating back to the 1960s[12–15], yet user rejection rates and dissatisfaction have remained consistently high[16–19]. Furthermore, in almost all tasks, myoelectric prostheses are outperformed by dramatically simpler body-powered prostheses[15,20,21]. This is because body-powered prostheses are operated continuously with minimal latency and provide feedback to the user[22–24], in what is known as closed-loop continuous control (CLCC). Achieving CLCC of multiple DOFs of myoelectric prostheses may be a large step towards restoring dexterity to users.

This is a non-trivial challenge, and previous research is often limited to attempting one half of the goal; either closing the feedback loop or achieving continuous control in isolation. To overcome poor EMG signal quality and lower sensory acuity associated with the skin of a user's residual limb, some works resort to invasive surgical procedures to either implant electrodes for EMG detection[25–28] or nerve stimulation[26,29–31], or for reinnervation of neural pathways[32]. While these studies show promising results, controlling up to 6 DOF[26], as case studies of one[26] or two[29,32] participants they are limited in how applicable they may be for a wider population. The uptake of such methods is currently restricted due to the specialised and experimental nature of its associated surgical procedures, and procedures such as targeted muscle reinnervation have, to date, not demonstrated reduced prosthesis abandonment in users[33]. In the most extreme case, with implanted electrodes, there is a risk that the small population of prosthesis users become reliant on commercial technology where there is no guaranteed long-term support or protection from obsolescence. Non-invasive techniques, utilising electrodes and feedback devices mounted on the skin's surface, may therefore present the most feasible route to closed-loop continuous control reaching a wider population. However, results of these techniques are much more limited, controlling fewer DOFs[34–36] and rendering few modalities of feedback to fewer feedback sites[37–40]. Existing works exploring CLCC using non-invasive technology have presented a single controlled DOF and single modality of feedback[41], or even just a single finger[42]. One of the most advanced examples of non-invasive CLCC comes from the work of Abd et al., who demonstrated CLCC of 2 DOF of a robot arm-mounted, sensor-equipped robot hand[40]. However, the applicability of this technique to real prostheses is questionable due to the robot arm-mounted hand, avoiding EMG data collection difficulties of a worn prosthesis, and due to sensory feedback being rendered to the user upon known object breakage, which requires a level of knowledge about the grasped object that is unobtainable in the real world. Furthermore, studies utilising tactile feedback[40,41] are limited in their application to currently used prostheses by the wider population, where tactile sensors are not commonplace, and are therefore limited to sensing only proprioceptive information. As such, there is significant scope for further research into non-invasive techniques to achieve proprioceptive CLCC of myoelectric prosthetic hands.

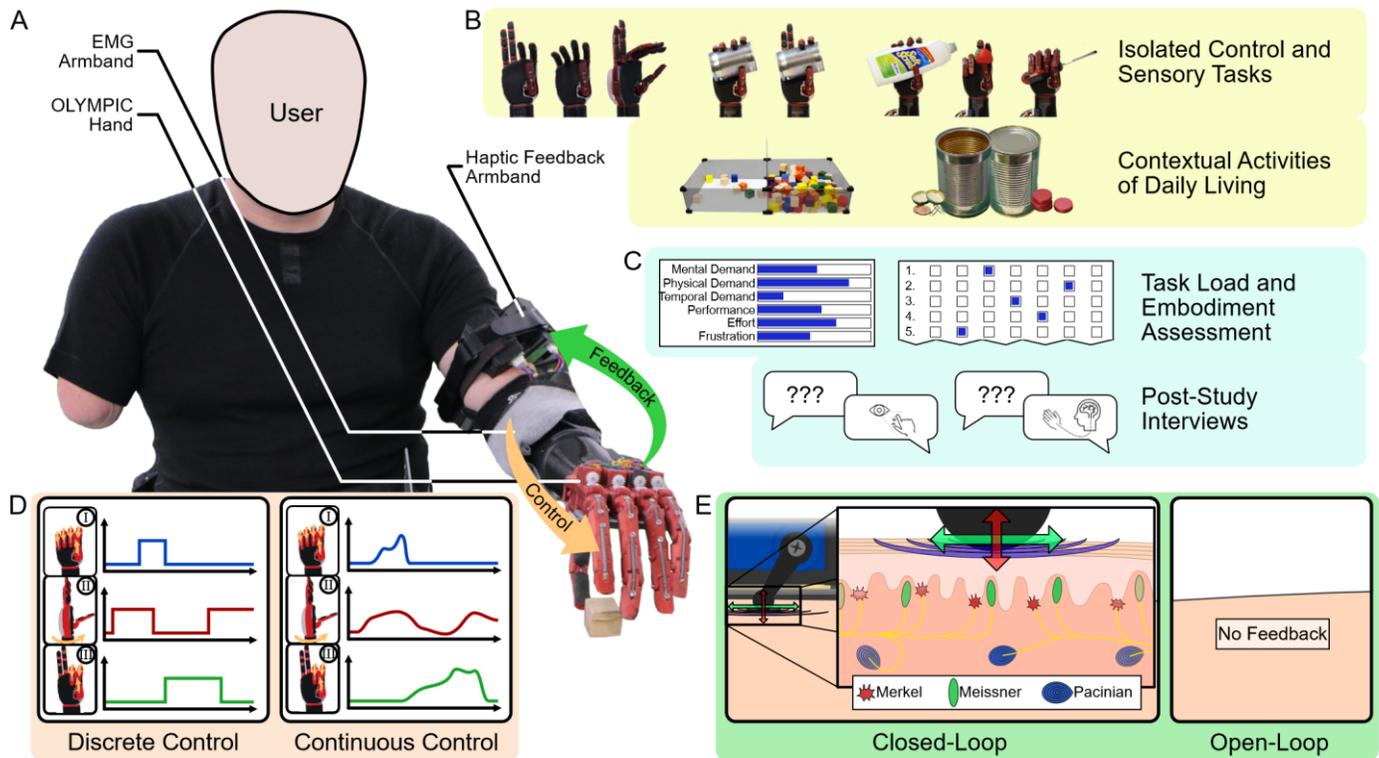

**Figure 1.** Non-invasive closed-loop continuous control achieved with surface electrodes and a haptic feedback armband. (A) A participant operating the OLYMPIC hand with the closed-loop continuous controller. (B) Physical performance evaluations. Isolated control tasks: position matching, force matching. Isolated sensory tasks: object identification. Contextual activities of daily living: Box and Blocks Test, Jebsen-Taylor Hand Function Test. (C) Psychological performance evaluations. Task load and psychological embodiment questionnaires, and post-study interviews. (D) Discrete Control and Continuous Control conditions. Discrete Control: reference actions are activated with binary signals. Continuous Control: reference actions are activated with modulated signals proportionally and simultaneously. (E) Open-Loop and Closed-Loop conditions. Open-Loop: no feedback is rendered to the user. Closed-Loop: multimodal proprioceptive feedback is rendered to the user with a haptic feedback armband.

Evaluation is central in prosthetic technology; prosthetic hand control is a complex, multifaceted relationship between human and technology. Both physical and psychological factors must be considered to obtain a more complete understanding of this. Often, invasive and non-invasive studies alike evaluate a small subset of these factors, such as isolated control and sensory performance[35,43,44], physical task performance[26,34], or user perception[45,46]. A 'holistic' study examining prosthetic hand control through multiple lenses could benefit wider scientific understanding.

In this study, we extend the state of the art in non-invasive methods with a novel closed-loop continuous prosthetic hand controller that can continuously modulate the commanded positions of three controllable DOFs of the OLYMPIC hand[4]: power grip (DOF I), wrist pronation/supination (DOF II), and tripod grip (DOF III), where grasping DOFs and wrist pronation/supination can be controlled simultaneously. We utilise a modified version a previously developed haptic armband[47], which renders complete proprioceptive feedback of the position and force of the motor driving each finger, and the motor driving wrist position to the user via 15 actuators, while maintaining a wearable form. Fig. 1 summarises this controller. 28 participants without ULD formed four groups: open-loop discrete control (OLDC), closed-loop discrete control (CLDC), open-loop continuous control (OLCC), and closed-loop continuous control (CLCC). As case studies, 10 participants with ULD used the CLCC setting. The controller was evaluated holistically: participants completed isolated control and sensory tasks, contextual dexterity assessments, and questionnaires on task load and embodiment. Finally, each participant with ULD completed a post-study interview to better understand their questionnaire results.

## Results

### Isolated Control Tasks

Position and force matching tasks evaluated the isolated control ability of participants. Participants matched 10 target positions for five single and dual combinations of controllable DOFs, and target forces for the two graspable DOFs, giving a total of 30, 20 samples per participant for single and dual DOF position matching tasks, respectively, and 20 samples per participant for grasping tasks. Mean absolute error (MAE) results for the OLCC and CLCC groups and participants with ULD are shown in Fig. 2. For position matching (Fig. 2 A-C), statistical significance was calculated between the OLCC and CLCC groups with a two-sided Mann-Whitney U test. During testing, the OLCC group exhibited a significantly higher absolute error than the CLCC group when performing dual DOF motions ($p < 0.001$), corresponding to a median absolute position error that was 54.0% greater than the CLCC group. Statistical significance was computed using a two-sided Mann-Whitney U test after Bonferroni correction for $N = 10$ comparisons between individual participants with ULD and the CLCC group. In four cases, individual participants with upper limb difference significantly outperformed the CLCC group on dual DOF motions (all $p < 0.0001$), corresponding to participants 001, 005, 008, and 009, who all were limited to only controlling DOFs I and II.

Force matching results, shown in Fig. 2 D-F, show that the OLCC group produced a signficantly higher absolute motor force error ($p < 0.05$, computed via a two-sided Mann-Whitney U test), corresponding to a median absolute force error that was 22.3% greater than the CLCC group. As in the position matching task, participants with ULD who were limited in their control capacity to only two DOFs — participants 001 ($p < 0.005$), 005 ($p < 0.001$), 008 ($p < 0.005$), and 009 ($p < 0.0001$) — performed significantly better than the CLCC group, computed with a two-sided Mann-Whitney U test after Bonferroni correction for $N = 10$ comparisons. Raw position and force error results of all four participant groups and individual participants with ULD are shown in Supplementary Figs. S17-S18.

### Isolated Sensory Tasks

Participants attempted to identify 5 objects while blindfolded; a bleach bottle, a plastic strawberry, a soft brick, a screwdriver, and a tuna can. These were presented 25 times in a randomised order (5 trials per object), after 2 training (non-blindfolded) trials. Results, shown in Fig. 3, show that the OLDC group were able to identify the bleach bottle significantly more accurately than the OLCC group ($p < 0.0083$, computed with a two-tailed Mann-Whitney U test after Bonferroni correction for $N = 6$ comparisons between each participant group), indicating that consistent auditory feedback from the motors of the prosthetic hand during constant speed grasping achieved during discrete control may play a driving role in object identification performance. The CLCC group performed almost identically to the OLDC group across each object, potentially indicating that the proprioceptive feedback provided by the armband was able to make up for the deficit in consistent auditory feedback during continuous control.

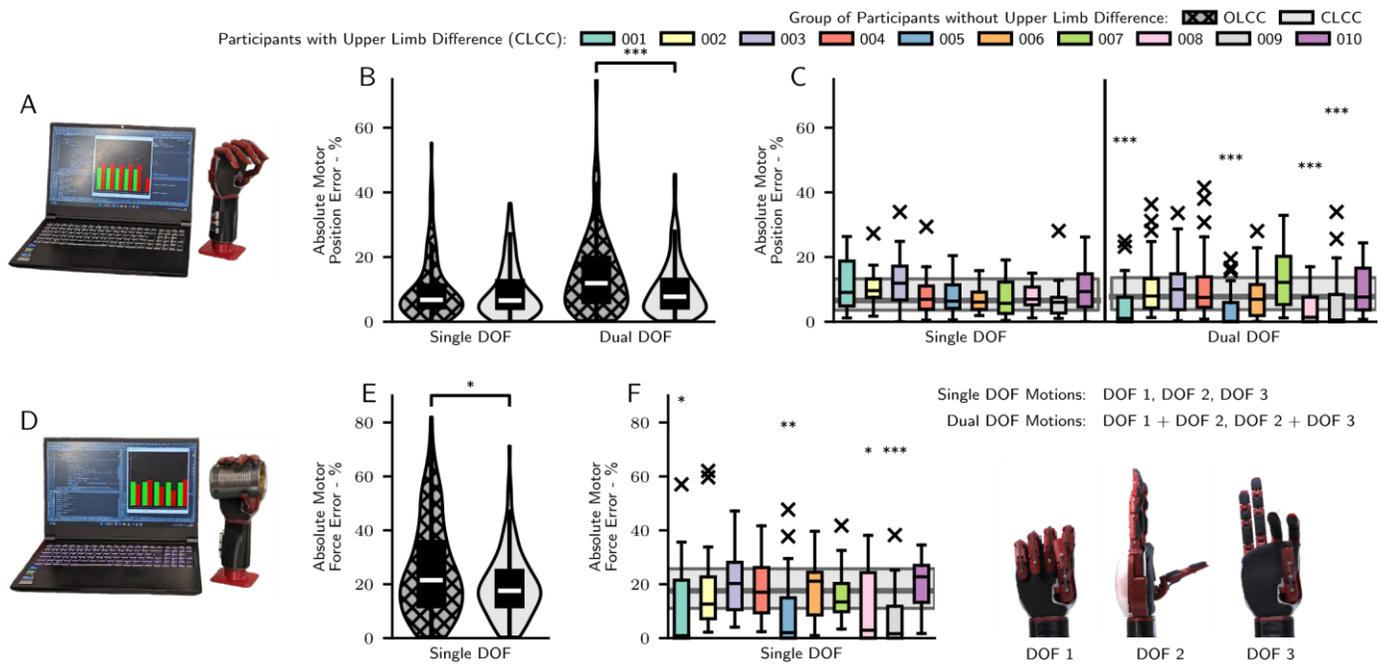

**Figure 2.** Participant results of isolated control tasks. Position matching results: (A) experimental setup of the position matching task, (B) absolute motor position error of groups of participants without upper limb difference, (C) absolute motor position error of individuals with upper limb difference, with median and interquartile range of the CLCC group without upper limb difference for single and dual DOF motions shown faded behind individual results. Force matching results: (A) experimental setup of the force matching task, (B) absolute motor force error of groups of participants without upper limb difference, (C) absolute motor force error of individuals with upper limb difference, with median and interquartile range of the CLCC group without upper limb difference shown faded behind individual results. Results of paired Mann-Whitney U tests at $*p < 0.05$, $**p < 0.01$, and $***p < 0.001$ levels between the OLCC and CLCC participant groups are shown as linked asterisks on (B) and (E). Results of paired Mann-Whitney U tests at $*p < 0.005$, $**p < 0.001$, and $***p < 0.0001$ levels after Bonferroni correction for $N = 10$ comparisons between individual participants with limb difference and the CLCC participant group are shown as floating asterisks on (C) and (F).

Participants with ULD generally performed similarly to the CLCC group in terms of average accuracy, which was equal to 51.6%±4.6% mean±SE on average across all participants with ULD, and 54.9%±4.9% mean±SE for the CLCC group. There were some exceptions: individual confusion matrices (shown in Supplementary Fig. S20) show that participants 002 and 003 struggled to distinguish between smaller diameter objects (the plastic strawberry, soft brick, and screwdriver). Participants 004 and 008 were unable to identify objects due to sensory impairment from skin graft and nerve damage, respectively, and even though both could perceive general changes in stimulation during isolated control tasks, they could not resolve this at finger-level to identify objects. Participant 006 consistently misidentified the bleach bottle as the soft brick and vice-versa — swapping classifications of these objects yields the faded results shown, which are closer to the expected CLCC group results.

**Contextual Activities of Daily Living**
The contextual performance of participants was evaluated with assembled versions of two clinical dexterity assessments: the Box and Blocks Test[48] (BBT) and the Jebsen-Taylor Hand Function Test[49] (JTHFT). Seen in Fig. 4, tasks involving static grasp show no difference between any group; this is expected when no different control strategy is needed. Task performance across the non-static tasks of the JTHFT was comparable across all groups, with a linear mixed effects model finding no significance between groups. BBT results were also comparable for the four participant groups, with no statistical significance between them.

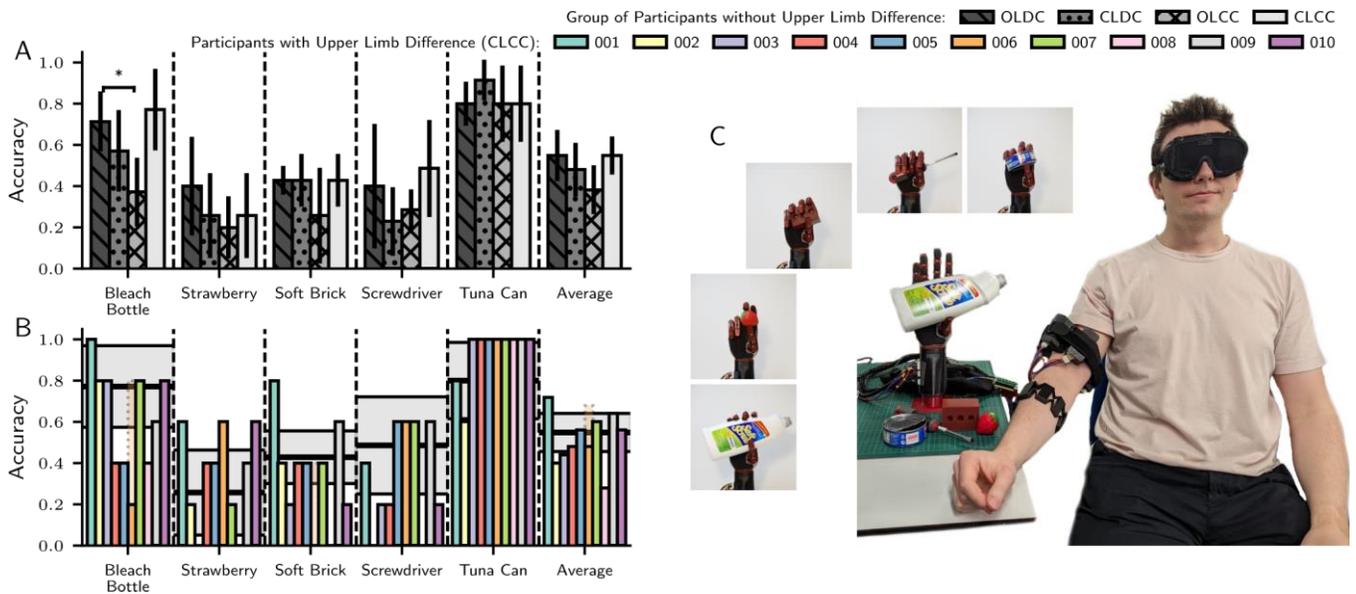

**Figure 3.** Participant results of isolated sensory tasks. (A) Object classification accuracy of groups of participants without limb difference. Statistical significance of paired Mann-Whitney U tests between each participant group shown at *$p < 0.0083$ levels after Bonferroni correction for $N = 6$ comparisons. (B): Object classification accuracy of individual participants with limb difference, mean and ±1 standard deviation of the CLCC group results are shown for comparison as a black line and shaded region, respectively. The faded results shown for participant 006 are produced by swapping their predictions for the bleach bottle and soft brick due to consistent misclassification between the two objects. Note: random chance is 0.2. (C) Experimental setup; participants grasped five objects five times in a random order without visual feedback. Objects (clockwise): bleach bottle, strawberry, soft brick, screwdriver, and tuna can. Confusion matrices of participant groups and individual participants with limb difference are shown in Supplementary Fig. S19 and S20, respectively.

Results of participants with ULD generally align with CLCC group results, with some notable exceptions. Participant 002 managed to 'hack' the simulated page turning task by flicking the paper over with the thumb extended. Participants 004 and 008 completed the small objects task much slower than other participants with ULD and the CLCC group, due to the condition of their ULD preventing compensatory movements from being performed. Participant 002 could not complete the small objects task and participants with ULD 001 and 002 could not complete the large, heavy objects task due to fatigue. The residual limbs of participants 001 and 010 with ULD were both such that neither participant had a functioning elbow joint, which prevented the completion of the small objects task. Participant 008 used an elastic sling held by their opposing arm due to partial paralysis of the right arm (see Supplementary Fig. S13h), which restricted wrist rotation and made the large, heavy objects task uncompletable. A linear mixed effects model was used to compare the CLCC group of participants without ULD and all participants with ULD, with no significance found.

**User Perception**

Participants completed task load and embodiment questionnaires four times during the study: stage 0 - pre-start, stage 1 - after position matching, stage 2 - after force matching and object detection, and stage 3 - after dexterity assessments. Task load was evaluated using the NASA Task Load Index (NASA-TLX) questionnaire[50], where aspects of task load are self-reported on a 20-point scale (displayed in Fig. 5 with a maximum value of 100, interval of 5). Fig. 5 shows the responses for physical demand, performance, and frustration (full questionnaire responses are shown in Supplementary Fig. S21). As seen, completing stage 1 with a continuous controller elicited a higher physical demand than for participants using a discrete controller, with statistical significance calculated using a linear mixed effects model between the groups using a continuous controller and the groups using a discrete controller ($p < 0.001$), and the CLDC group ($p < 0.05$). This is expected — discrete control users could only produce binary motions and forces, and therefore were not required to use the prosthesis to a point where physical demand becomes high. After stage 1, this effect is negated for the CLCC group, who have a 20.2%, 22.7%, and 31.2% lower reported physical demand for stage 3 than the OLDC, CLDC, and OLCC groups, respectively. Interestingly, even with the added 0.40 kg of the haptic feedback armband, closed-loop groups show no associated increase in physical demand, potentially because it is mounted close to the body on the upper arm. Perceived performance is approximately consistent across groups for stages 1 and 3, whereas in stage 2 the CLCC group exhibit a significantly higher perceived performance than the CLDC group ($p < 0.0083$ after Bonferroni correction for $N = 6$ comparisons) found using a linear mixed effects model. Compared to the OLDC, CLDC, and OLCC groups, this corresponds to a 45.9%, 56.5%, and 31.7% relative increase in respective perceived performance by the CLCC group. This is expected, given that success in the force matching aspect of stage 2 required both continuous control and sensory feedback. Perceived frustration of participant groups show no statistical significance between groups, however, by stage 3, CLCC group frustration is 36.4%, 41.7%, and 48.1% lower than the OLDC, CLDC, and OLCC groups, respectively. While participant groups have relatively low variance, participants with ULD are extremely heterogeneous. Physical demand for participants with ULD was higher than those without across all stages. The majority of participants were not prosthesis users, and those that were used body-powered or static prostheses. Activating muscles that are rarely used and a larger weight on the residual limb heightens the physical toll of myoelectric prosthesis use. For many participants, frustration was linked to perceived performance. This is most visible for participants 002, 004, 005, and 008 at stage 2.

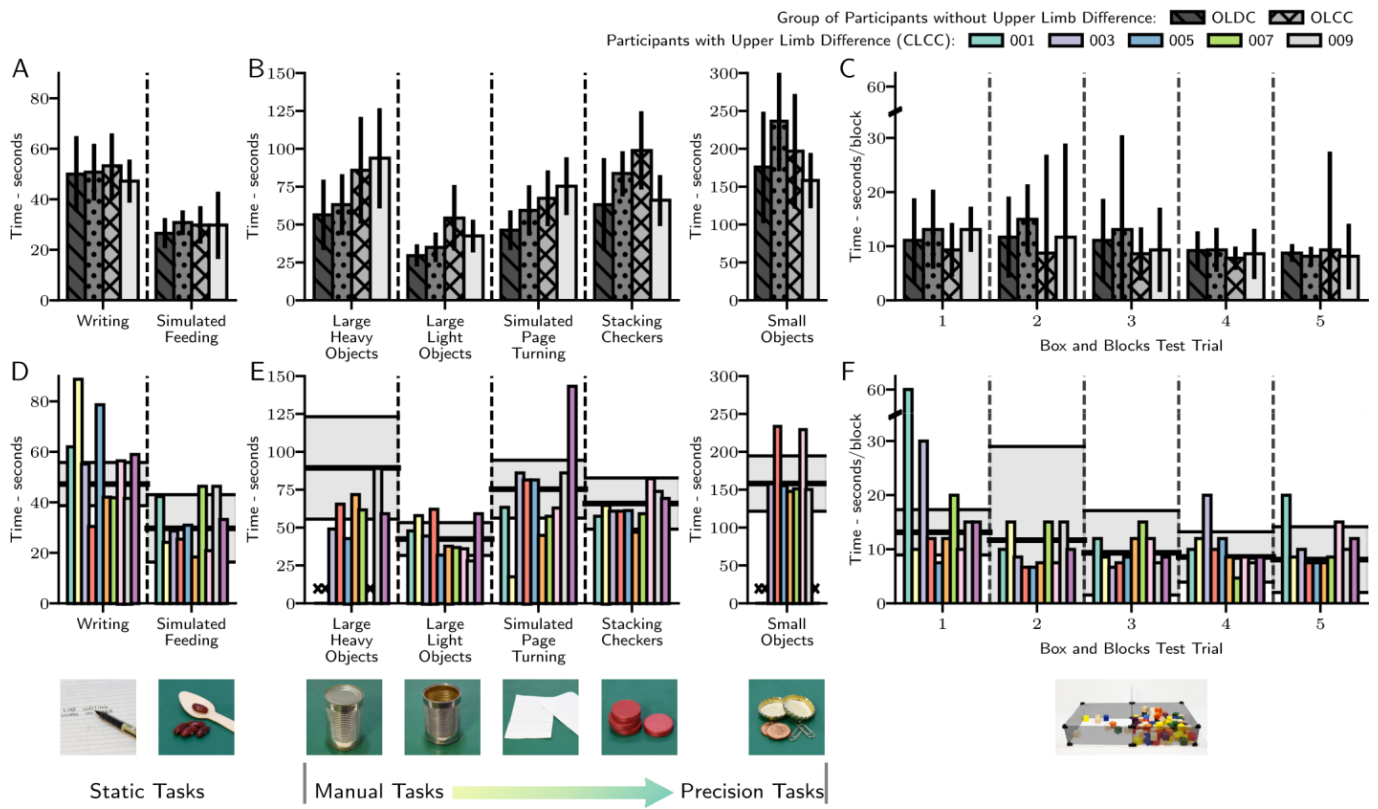

**Figure 4.** Participant results of dexterity assessments. (A - H) Dexterity assessment results of groups of participants without limb difference (A-C) and individual paticipants with limb difference using the closed-loop continuous controller (D-F) performing tasks from the Jebsen-Taylor Hand Function Test (A and B, D and E), and the Box and Blocks Test (C, F). In (D-F), mean and ±1 standard deviation of the CLCC group results are shown for comparison as a black line and shaded region, respectively.

Fig. 5 also shows the responses of participant groups and individual participants with ULD to statements from the embodiment questionnaire. A post-study interview was conducted with each participant with ULD to further understand perception responses. Two primary perception-influencing factors for each participant with ULD were identified from interviews, and are shown along with a summary of participant characteristics in Table 1.

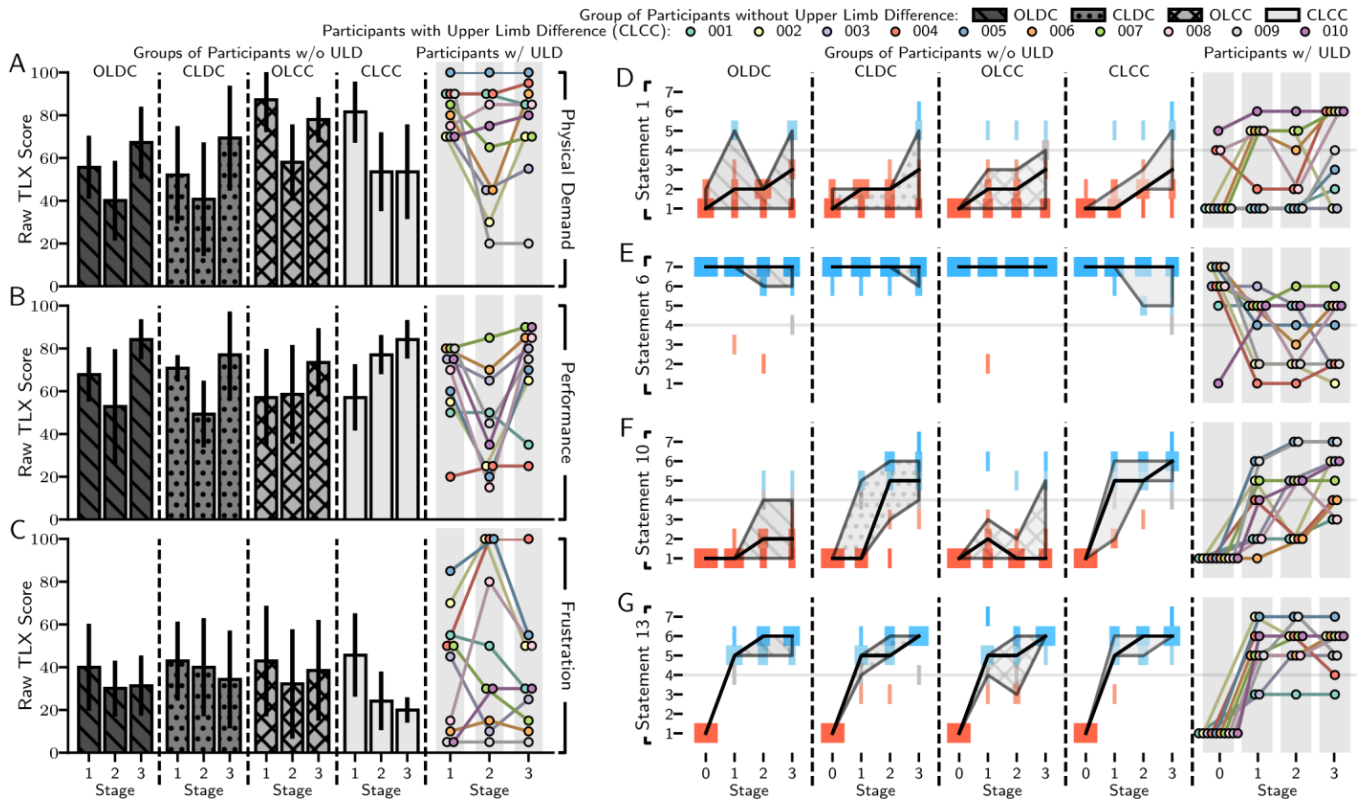

**Figure 5.** Task load and prosthetic embodiment responses of participants. (A-C) Responses of participants to the NASA-TLX questionnaire: (A) physical demand, (B) perceived performance, and (C) frustration. (D-G) Responses of participants to the prosthetic embodiment questionnaire statements 1. 'The prosthesis is my hand' (D), 6. 'My body feels complete' (E), 10. 'I could feel the position of the prosthesis' (F), and 13. 'I am in control of the prosthesis' (G). Median response and the interquartile range of groups of participants without limb difference to the embodiment questionnaire are shown as a solid black line and shaded region, respectively. Response values are as follows: 1. Strongly Disagree, 2. Disagree, 3. Somewhat Disagree, 4. Neutral, 5. Somewhat Agree, 6. Agree, 7. Strongly Agree.

Ownership (statement 1) increased for most participants, with the sharpest rise coinciding with donning the prosthesis in stage 3. In contrast to this general trend, participant 002 reported a sharp drop in ownership stage 3, even though perceived performance, volition, and vividness were all relatively positive. Here, a high value was placed on a lightweight prosthesis and secure socket, which could not be achieved with the research socket. During the desk-mounted stages 1 and 2, embodiment was partially aspirational: "...when it's on the table, you can see it's a hand and you can relate to that as being your hand [...] with the hope that it will work and you will gel with it". Participant 008 exhibited a decreased ownership after stage 2; visuomotor connection with the prosthesis was important in incorporating it into their body, made even more important by nerve damage heavily impairing acute sensory feedback from the haptic armband. During stage 3 this visuomotor connection was restored, and ownership increased again.

Participants without ULD consistently responded positively to body completeness (statement 6), in contrast to participants with ULD, whose responses actually became significantly more negative over the course of the study ($p < 0.01$). This was a significant difference in embodiment questionnaire response ($p < 0.001$) between participants with and without ULD, who otherwise responded comparably to every statement. This was prominent for participant 002, a non-user in their daily life, who dropped from 'Strongly Agree' to 'Strongly Disagree' from stages 0 to 3, stating that when "dealing with prostheses it comes very quickly back to you that you have lost an appendage". Completeness fluctuated for participants 006 and 008: both experienced a drop at stage 2 which was restored at stage 3, aligning with the removal of the visual connection with the prosthesis during stage 2. Participants 007 and 010 experienced a positive trend in completeness. Participant 007, who had a congenital ULD, responded 'Agree' at stage 0, which dropped slightly to 'Somewhat Agree' after stage 1, then increased to 'Agree' after stage 2, where it remained. Participant 010, a relatively recent patient of amputation, responded 'Strongly

**Table 1.** Summary table of participant characteristics of participants with limb difference and identified factors that influence perception.

| Participant | ULD | Controlled DOFs | Cause | Years Since | Own Prosthesis | Frequency of Use | Factor 1 | Factor 2 |
|---|---|---|---|---|---|---|---|---|
| 001 | RABE* | 2 | C (TI) | 26 (8) | Cosmetic | Infrequent | Functionality | Aesthetics |
| 002 | LABE | 3 | PS | 5 | None | - | Comfort | Functionality |
| 003 | LABE | 3 | C | 30 | None | - | Aesthetics | Functionality |
| 004 | (RABE) LAAE | 3 | TI | 9 | Body-Powered | Daily | Functionality | Comfort |
| 005 | RABE | 2 | C | 31 | Static | Infrequent | Visuomotor | Functionality |
| 006 | RABE | 2 | C | 30 | None | - | Visuomotor | Aesthetics |
| 007 | LABE | 3 | C | 28 | Static | Frequent | Functionality | Aesthetics |
| 008 | RABE† | 2 | TI (PS) | 24 (6) | Static | Infrequent | Visuomotor | Aesthetics |
| 009 | LABE | 2 | C | 22 | Static | Frequent | Functionality | Aesthetics |
| 010 | LABE* | 3 | TI | 3 | Static | Infrequent | Visuomotor | Functionality |

Upper limb difference sides listed: RAXX (right arm), LAXX (left arm); levels listed: XXBE (below elbow), XXAE (above elbow). For bilateral upper limb differences the side tested is shown in brackets, and details for the tested side are shown. Causes listed: congenital (C), traumatic injury (TI), planned surgery (PS). For participants with secondary causes of upper limb differences, secondary cause is shown in brackets. Factors listed are obtained from the post-study interview. *Participants 001 and 010 had high limb differences, resulting in no functioning elbow joint. †Participant 008 had partial paralysis of the right arm and partial muscle reinnervation following nerve transplant, meaning EMG signals were collected from the reinnervated region of their upper arm. Full participant characteristics are listed in Supplementary Table S2.

Disagree' at stage 0, which increased to 'Somewhat Agree' after stage 1, where it remained. Both participants cited functionality as a key factor, with participant 007 stating "I was less reliant on my right hand, and actually having another option led to me feeling more complete".

Sensory feedback, predictably, led to a significantly more positive response to statements 9, 10, and 11, which assessed perceived touch, position, and force in groups with sensory feedback compared to groups without sensory feedback ($p < 0.001$ for statements 9, 10, and 11). This is somewhat surprising when considering that the feedback rendered to participants was proprioceptive, so tactile feedback was not directly included.

Control (statement 13) became increasingly positive throughout the study, primarily driven by exposure to each controller. Participants with ULD responded similarly: an increase in perceived control for participant 001 corresponded with an improved understanding of the combined control and sensory feedback, a sentiment shared by participants 007 and 008. Similarly, participant 005 perceived control becoming more natural as tasks became more contextual, "By the end it wasn't 'contract this muscle', it was 'pick that up'. Then I was doing it as a natural reaction". Aspirational use was important to participant 007, and simultaneously actuating DOFs aided this, "grabbing something and turning, I could see how that would be useful". Participant 003 responded similarly, however was limited by the study environment, "The exercises felt like not 'real' experiences in that we were [...] in a research lab. They're not tasks that I'd be doing in my day-to-day life... What would make me feel like it was my hand would be to use it in real life scenarios". Continuous control was beneficial for participant 008, "thinking of that action was sort of transported [...] into the hand in a smoother movement, whereas if it was jerky, I would never have been able to [perform precision tasks]".

Participant responses to all embodiment questionnaire statements are shown in Supplementary Fig. S22.

## Discussion

*Benefits of Closed-Loop Continuous Control*

Results demonstrate that the CLCC group outperform other groups in isolated control tasks, and on par with conventional open-loop discrete control on the isolated sensory task and dexterity assessments, and that individual participants with ULD generally demonstrate performance levels comparable to those without ULD using the same controller. Exceptions to this were generally linked to the physical condition of each participant's upper arm or residual limb.

Group responses to the task load questionnaires highlight the benefit of closed-loop continuous control when performing tasks that require continuous control and sensory feedback (such as force matching and object identification in stage 2). Participants with ULD responded similarly in terms of task load to those without ULD, however a larger variance in participant responses was noted, potentially indicating that individuals with ULD are more heterogeneous and at risk of negative response. Embodiment results highlight that sensory feedback promotes a perception of touch, position, and force. Responses of participants with ULD fell within expected ranges for the CLCC group for the majority of embodiment questions, showing that these benefits can translate to real users with ULD.

Although results are positive, the introduced controller falls short of participant desires in some areas. The most common requests were reduced delay and improved accuracy. Delay occurs because the controller is reactive, rather than predictive, and because of latency in the prosthesis. Accuracy issues, across all groups, occurred when carrying out tasks while wearing the prosthesis. The added weight on the residual limb is the main cause of this, and can be mitigated by recalibration and user practice to isolate muscle groups.

*Open-Loop Discrete Control Provides Key Benefits*

The effectiveness of standard open-loop discrete control in many scenarios is difficult to ignore. This was most prominent in object identification, where the OLDC group were on par with the CLDC group, based on auditory feedback alone. Audition as a source of 'incidental' feedback, where the sound of finger motors can be used to identify motor position and grasping force from the duration and volume of auditory feedback, respectively, among other things[51,52], is a well-known skill obtained by prosthetic hand users. This ability is reduced when using a continuous controller because motor speed is no longer consistent, resulting in poor identification performance of the OLCC group. Feedback did not improve object identification performance of discrete control users. During grasping, the dominant feedback mechanism was normal displacement, rendering finger torque. Because the hand is closed discretely with maximum force, there is little differentiating the feedback from each finger to a participant. Instead, the transient feedback of sequential increases in finger torque as each finger contacts an object must be observed. It is possible that tactile sensors on the prosthesis would improve object identification and overcome proprioceptive feedback limitations, as well as promoting embodiment.

Discrete control groups also completed dexterity assessment tasks in similar times to the continuous control groups, possibly because discrete control only requires muscle activity to reach a desired pattern to fully close the hand. This is somewhat faster and easier to sustain than maintaining a fully closed hand with continuous control. This effect was most visible on the large, heavy objects task, in which DOF I and II must be fully activated together to cradle the heavy object in the palm of the prosthesis in order to prevent it slipping from the grasp. The benefits of discrete control can be leveraged in future continuous controllers. Methods such as mode-switching, popular in extending the functionality of conventional controllers, between discrete and continuous control could allow users to benefit from a full range of manual and precise dexterity.

*End User Heterogeneity Leads to Complex Perception Results*

It is easy for researchers to assume that participants who acquired an ULD later in life may feel 'less complete' or more 'in need' of a prosthesis than someone with a congenital ULD. We find that this is not necessarily the case. All participants with ULD in this study, except for participant 010, viewed their body completeness positively at stage 0, with no clear separation between participants with congenital ULD versus acquired ULD. Participant 004 went so far as to say, "I've had two bodies. I had one with arms and legs and now I've got a body with no legs, bit of an arm missing, a lot of the other arm missing. 'Body complete' to me is just wearing that one prosthesis on my right arm.". By contrast participant 010, who relatively recently had undergone amputation, stated "I know my body is not complete because I can look at [my arm], it's gone". During the study, varied results in terms of body incorporation and completeness were observed, confirming high variation reported in literature[53], however the cause of a participant's ULD was only a determining factor of perception when it was tied to their performance in a task. Body completeness was the only embodiment statement where a significant difference between participants with and without ULD was found, however longitudinal studies will be needed to identify other statements with diverging responses.

Ideally, enough participants with ULD would be recruited such that analysis is adequately-powered to observe statistical differences between intervention conditions and a control group. In reality, this is improbable due to a small population size, difficulty of recruitment, and significant participant heterogeneity. As a result, studies often rely on a small number of participants with ULD and supplement these with a larger sample of participants without ULD[5,15,54]. While such studies, including this one, often show similar physical performance results between participants with ULD and their counterparts without ULD, regardless of demographic differences, new technology often fails to make the leap from research to wider adoption. This, coupled with unchanging rates of rejection[19,55], make it clear that there is something missing in terms of researchers' understanding of prosthetic hand use from an end-user perspective. Individual participants with ULD with similar physical circumstances and task results can respond very differently, psychologically, to prosthetic hand use. When a small number of participants with ULD are recruited, effort should be made to holistically understand each participant's experience. User expectations and priorities play a vital role in the developing relationship with their prosthesis. As summarised in Table 1, the aesthetic qualities of the prosthesis were often an important—but secondary—factor. The anthropomorphic structure of the hand was praised, and played a role in promoting natural motion, ownership, and completeness. However, the length of the prosthesis in relation to their expectation of where their hand should be received mixed responses, with participant 003 stating that the socket extending above the elbow made the prosthesis appear longer than it physically was. Comfort was often a secondary factor in perception, potentially due to the prosthesis only being donned for stage 3, and also that participants were confident that a commercialised iteration of the device would be more comfortable. For others, comfort took on a central role; without a secure socket and lightweight hand, embodiment was more difficult to obtain. Functionality as a priority was often linked with prior non-anthropomorphic prosthesis use. For such participants, body incorporation remained low, even though ownership increased. For example, participants 004 and 009 finished the study with positive and neutral responses to "the prosthesis is my hand", but negative and strongly negative responses to "the prosthesis is part of my body", respectively. Corroborating this, participant 001 stated, "[my left] hand is a part of me, whereas I forever see anything [on my right] as something that is just an extension; artificial. So I assign value to being a part of me is how useful that tool is at the end". The visuomotor connection with a prosthesis can also be critical; the perception that a desired motor signal is related to the visible motion of the prosthesis is intricate, with many contributing aspects such as latency and the correspondence between residual muscle- and prosthesis-actuation. For participants where this was important, the controller's continuous motion promoted a perception of natural motion, in turn promoting embodiment. Familiarity plays a large role in embodiment[56], however, due to the short-term nature of this study, it was difficult for participants to build upon this. This is a drawback of this study, however, even within the time-frame, participants commented on a positive relationship between duration of use and embodiment, with participant 007 stating, "...as my familiarity with the prosthesis grew, it did make me feel like I was looking at more of a hand than a prosthesis".

Importantly, most participants with ULD in this study had previously rejected myoelectric prostheses. Identifying points when task load increased or embodiment decreased can provide clues to the reasons behind rejection. For example, understanding that participant 002's sharp drop in embodiment after donning the prosthesis was related to comfort would enable training to be adjusted to mitigate this. Moreover, rejection is a worst-case outcome; acceptance of the prosthesis by its user is the ideal scenario. Acceptance is difficult to monitor, and a holistic approach is needed over conventional outcome measures.

## Methods

### Prosthetic Hand Control

In all experiments, the OLYMPIC hand[4] was used as a test platform. Core to the discrete and continuous controllers is the estimation of the distribution of observed signals using kernel density estimation (KDE). That is, the probability of observing a signal with intensity $I$ at circumferential position $\vartheta$ around the forearm is estimated from rectified EMG signals x:

$$p(I, \vartheta) = KDE(x). \qquad (1)$$

As a baseline control method, a previously developed discrete controller[15] was used. Discrete control is achieved by computing the Wasserstein distance between the live distribution of muscle activity and a set of pre-recorded reference muscle activities, then selecting the closest reference that falls below a distance threshold. The reference motion corresponding to the reference muscle activity is then sent as a command to the prosthetic hand.

To extend this to perform as a continuous controller, instead of performing the discrimination task of selecting the reference with the minimal Wasserstein distance to the live distribution, we perform the generative task of approximating the live distribution from known references to minimise the Wasserstein distance between the live and approximate distributions. When multiple actions are performed with independent muscle groups that do not form antagonistic pairs (henceforth called complementary actions), the resultant muscle activity that is detected by an electrode is the superposition of the EMG signals produced by each of the muscle groups[57]. This concept is applied to the intensity dimension of the kernel density estimates of multiple reference muscle activities, producing samples that approximate incoming, live EMG data. The Wasserstein distance between the approximate and live data is used as a loss function by which to update the weighting of each reference via gradient descent for a fixed number of optimisation steps per control step.

To convert computed weights to an output motion, each reference is mapped to the control of a DOF of the prosthesis. Complementary references are mapped to independent motions. For example, the motion of the wrist and fingers. Antagonistic references are mapped to overlapping motions such that they cannot be activated at the same time. For example, tripod grip and power grip. The target position at control step $k$, $P_i$ of the prosthesis is the weighted sum of each reference, exponentially smoothed with its previous position:

$$P_k = \alpha \sum_{i=1}^{r} w_i P_i + (1-\alpha) P_{k-1}, \tag{2}$$

where $\alpha$ is the smoothing factor, chosen heuristically as 0.3. Three DOFs were selected for control: DOF I, a power grasp, DOF II, wrist pronation/supination, and DOF III, tripod grip, chosen to enable the user to perform manual and precision grasps[58], with the simultaneous secondary motion of rotating the hand. Reference muscle activity patterns for participants without ULD were selected such that the two grasping DOFs were activated with antagonistic motions, wrist flexion and extension, while DOF II was activated by performing wrist abduction. For participants with ULD, references were chosen according to their existing abilities. DOF I was always selected to be their most comfortable action, DOF II a complimentary action that could be performed simultaneously to DOF I, and, where possible, DOF III was chosen to be an antagonistic action to DOF I. Further details of the discrete and continuous controllers are given in Supplementary Information S1.

**Sensory Feedback**

Proprioceptive feedback is rendered to the user to represent a combination of the position and force aspects of proprioception of the fingers and wrist of the OLYMPIC hand, which is equipped with sensors to detect finger position and force[59]. A modified version of a previously developed haptic feedback armband[47] is used, rendering three modalities of feedback at each of the five modules. The modes of feedback, tangential position, normal displacement, and vibration, render proprioceptive finger position, finger force, and wrist position, respectively.

Tangential position rendered at each module is proportional to that module's corresponding finger motor position. Normal displacement is rendered proportional to the difference between the corresponding finger's motor torque and the motor torque required to overcome the extension springs mounted on that finger[60]. Wrist position is rendered sequentially by the modules such that each module vibrates with an intensity proportional to a Gaussian curve with a mean and variance such that each module activates in turn, and when one module is at maximum intensity, other modules are below 1% maximum intensity. Further details of the haptic feedback armband are given in Supplementary Information S2.

**Study Design**

A mixture of participants with and without ULD were recruited. Each participant was fit with the OLYMPIC hand using a commercial soft-shell socket (ALX sleeve, Koalaa Ltd.), a non-medical electrode armband (Myo Armband, Thalmic Labs) and fit with the haptic feedback armband on their upper arm. Participants without ULD formed four groups: open-loop discrete control (OLDC), closed-loop discrete control (CLDC), open-loop continuous control (OLCC), and closed-loop continuous control (CLCC), with characteristics shown in Supplementary Table S3. Participants with ULD all had an ULD at transradial level or below (see Supplementary Table S2), and used the closed-loop continuous controller. To gain a greater understanding of the physical and psychological factors that influence a user's relationship with a prosthesis we implement a holistic approach to evaluating participant outcomes. In terms of physical performance, the controller was first assessed in isolation, with tasks targeted at both the continuous and closed-loop aspects of the controller, then in context, with two clinical dexterity tests. Regarding psychological factors, task load incurred while using the prosthesis and psychological embodiment were assessed.

Over three sessions, participants completed position matching tasks (stage 1), force matching tasks and an object identification task (stage 2), as well as clinical dexterity assessments (stage 3). A questionnaire designed to assess psychological embodiment and task load was administered prior to participation and at the end of each session. Participants with ULD were also asked to complete a post-study interview, the transcripts of which are contained in Supplementary Information, Transcripts S1 - S10. During stages 1 and 2, the prosthesis was desk-mounted, to allow the controllers to be evaluated in isolation, while in stage 3, the prosthesis was donned, to evaluate the controllers during contextual use. This study was given a favourable opinion by the Imperial College Research Governance and Integrity Team (RGIT), number 22IC7796, which was in accordance with the declaration of Helsinki. Written consent was obtained from all participants prior to taking part in this study. Participants depicted in figures gave informed consent for their images to be published.

**Experimental Setup**

The ability of participants to perform isolated control tasks was evaluated with position matching and force matching tasks. For these tasks, at each trial users were shown a target position or grasping force of the hand, and were asked to control the prosthesis to match this reference. In training, users were able to see a visual readout of the position or force of each motor in the prosthesis (see Fig. 2, (B) and (E)) for 10 training trials, then this visual feedback was removed during 10 testing trials. Target position complexity was graduated, with 3 rounds of single DOF targets, and 2 rounds of 2 simultaneous DOF targets. Mean absolute error was measured across the motors involved in each DOF; the 5 finger motors for DOF I, the wrist motor for DOF II, and the thumb, index, and middle finger motors for DOF III. The physical setup of the position matching and force matching experiments, including user interface of the testing software, are shown in Supplementary Figs. S14 and S15, respectively. Due to the binary nature of discrete control, the discrete control groups could not match positions or forces, and therefore the results from these groups are not included in Fig. 2. Despite this, all participant groups did complete the position and force matching tasks to ensure that exposure and use time of the prostheses was comparable across all four groups for embodiment and task load comparisons.

Participant performance of isolated sensory tasks was evaluated with an object identification task. In this experiment, 5 objects from the YCB object and model set[61] were used to provide objects of varying dimension and stiffness. The objects chosen were: bleach bottle, plastic strawberry, soft brick, screwdriver, and tuna can. The bleach bottle and soft brick were both relatively large, with the bleach bottle being rigid, and the soft brick being compliant. The plastic strawberry and screwdriver had similar width, but were grasped by three fingers and five fingers, respectively. Finally, the tuna can was selected because it made an audible sound on contact with the fingers of the prosthesis; prosthetic hand users are well known to utilise incidental auditory feedback[51], and poor classification of this object may explain wider classification performance. During training, participants were asked to execute power grasps of each object 2 times with visual feedback. During testing, visual feedback was removed using a blindfold, and participants attempted to identify each object 5 times in random order by executing power grasps. The physical setup of the object matching experiment is shown in Supplementary Fig. S16.

The performance of participants when completing contextual activities of daily living was evaluated with assembled versions of two dexterity assessments: the Box and Blocks Test (BBT)[48] and the Jebsen-Taylor Hand Function Test (JTHFT)[49]. Both dexterity assessments are widely used in clinical practice, and report task completion time as their performance metric. The BBT was completed 5 times, and participants were allowed 2 practice attempts at each JTHFT task before testing, to negate the effects of task-based learning. The physical setup of each dexterity assessment is described in Supplementary Section S3. Statistical analysis of dexterity assessments was performed using linear mixed effects models, fitted using SPSS statistical analysis software (IBM; UK).

To assess task load and embodiment, participants completed a questionnaire (listed in Supplementary Information S4) at the beginning, and after each stage of the study. The NASA Task Load Index (TLX) questionnaire[50] was used, and the raw TLX scores[62] presented, to quantify perceived level of six comprising factors of task load: mental demand, physical demand, temporal demand, (perceived) performance, effort, and frustration. Of relevance to prosthetic hands, due to their parallels with commonly cited reasons for rejection[63], are physical demand, related to prosthesis weight and fatigue while using a prosthesis, perceived performance, related to perceived functionality, and frustration. Psychological embodiment of the prosthesis was assessed using a 7-point Likert-response questionnaire based on the Prosthesis Embodiment Scale for Lower Limb Amputees[46] and the Avatar Embodiment Questionnaire[64] (which are in turn based on the rubber hand illusion[65]) with two added questions to gain understanding of the role of proprioceptive feedback in embodiment. Statistical analysis of task load and embodiment questionnaire responses was performed using linear mixed effects models, fitted using SPSS statistical analysis software (IBM; UK).

Finally, to obtain a more in-depth understanding of the responses of participants with ULD to each questionnaire, a semi-structured post-study interview was conducted with each individual participant with ULD. Transcripts of these interviews are available in Supplementary Transcripts S1 - S10.

## Data Availability

The raw data and material used and analyzed in this study are available from the corresponding author upon reasonable request.

## Acknowledgements

The authors would like to thank all participants for the time given to take part in this study. The authors would like to thank the Alex Lewis Trust and Koalaa Ltd. for their support in development of the prosthetic hand and socket, and the UKRI Centre for Doctoral Training in Artificial Intelligence for Medical Diagnosis and Care (University of Leeds). The authors also thank Allan Pang, David Hogg, Dongmyoung Lee, William Duggleby, Heval Turel, Oliver Meysner, Jordan Shillings, Ruth Tomlinson-Clark, Alexandra Reece, Simon Reece, and Alison Reece for assistance in facilitating research, Yukun Ge for artistic services, and Claire Traweek, Harrison Young, Peter Kyberd, Barry Mulvey, Thilina Lalitharatne, and Thrishantha Nanayakkara for meaningful feedback on the manuscript.

## Author contributions

D.C., W.B., F.B., P.K., and N.R. designed research; D.C., Z.Y., and A.C. performed research; D.C., Z.Y., A.C., A.B., and C.L. contributed materials; D.C. analyzed data; and D.C., Z.Y., A.C., A.B., C.L., W.B., F.B., P.K., and N.R. wrote the paper.

## Funding

This work was supported by the UKRI CDT in AI for Healthcare (Grant No. EP/S023283/1).

## Competing interests

The authors declare no competing interests.